\documentclass{llncs}
\usepackage{makeidx}  
\usepackage{graphicx}
\usepackage{amsmath}
\usepackage{gensymb}
\usepackage{amssymb}  
\usepackage{algorithm,algorithmic}
\usepackage{bm}
\usepackage{multirow}
\usepackage{nicefrac}
\usepackage{textcomp}

\usepackage[pdftex]{hyperref}

\begin{document}

\mainmatter              
\title{A Comprehensive Approach for Learning-based Fully-Automated
Inter-slice Motion Correction for Short-Axis Cine Cardiac MR Image Stacks}
\author{Giacomo Tarroni\inst{1} \and Ozan Oktay\inst{1} \and Matthew Sinclair\inst{1} \and Wenjia Bai\inst{1} \and Andreas Schuh\inst{1} \and Hideaki Suzuki\inst{2} \and Antonio de Marvao\inst{3} \and Declan	O'Regan\inst{3} \and Stuart	Cook\inst{3} \and Daniel Rueckert\inst{1}
\\
\authorrunning{G. Tarroni et al.}}

\institute{BioMedIA Group, Department of Computing,
  Imperial College London, UK
  \and
  Division of Brain Sciences, Department of Medicine,
  Imperial College London, UK
  \and
  MRC London Institute of Medical Sciences,
  Imperial College London, UK
}

\maketitle              

\begin{abstract}

In the clinical routine, short axis (SA) cine cardiac MR (CMR) image stacks are acquired during multiple subsequent breath-holds. If the patient cannot consistently hold the breath at the same position, the acquired image stack will be affected by inter-slice respiratory motion and will not correctly represent the cardiac volume, introducing potential errors in the following analyses and visualisations. We propose an approach to automatically correct inter-slice respiratory motion in SA CMR image stacks. Our approach makes use of probabilistic segmentation maps (PSMs) of the left ventricular (LV) cavity generated with decision forests. PSMs are generated for each slice of the SA stack and rigidly registered in-plane to a target PSM. If long axis (LA) images are available, PSMs are generated for them and combined to create the target PSM; if not, the target PSM is produced from the same stack using a 3D model trained from motion-free stacks. The proposed approach was tested on a dataset of SA stacks acquired from 24 healthy subjects (for which anatomical 3D cardiac images were also available as reference) and compared to two techniques which use LA intensity images and LA segmentations as targets, respectively. The results show the accuracy and robustness of the proposed approach in motion compensation.

\end{abstract}
\section{Introduction}
Cardiovascular magnetic resonance (CMR) imaging is the reference technique regarding several applications for the anatomical and functional assessment of the heart \cite{Zhuang2011a}. While fast SSFP sequences allow the direct acquisition of an anatomical 3D image (A3D) of the whole heart, they are usually limited by either relatively poor image quality or low temporal resolution, making them often unsuitable for accurate functional assessment. The most common CMR sequence currently used in the clinical practice is still the short axis (SA) SSFP cine, consisting of 10-14 parallel slices and 20-30 frames per cardiac cycle. SA cine stacks are generated during multiple breath-holds (i.e. 1-3 slices acquired per each breath-hold). Although the subjects are instructed to hold their breath at the same breath-hold position, in practice the heart location can vary considerably. If the differences between the breath-hold positions are large, the acquired image stack will be affected by inter-slice motion and not correctly represent the cardiac volume, introducing potential errors in the following analyses and visualisations.

\noindent \textbf{Related Work.} Several approaches for SA stack motion correction (MC) have been proposed in the literature. Among the techniques that make use of routinely acquired CMR images, Lotjonen et al. \cite{Lotjonen2004a} proposed to perform in-plane rigid registration of each SA slice to LA images, used as target. Sinclair et al. \cite{Sinclair2017} implemented a similar approach using LV segmentations (obtained using a fully-convolutional neural network, FCN) instead of the actual images. A very similar technique was also developed by Yang et al. \cite{Yang2017}, which also included a shape model to better retrieve the actual motion of the myocardium throughout the cardiac cycle. An alternative approach, which has the advantage of being applicable even if LA images are not available, consists in implicitly incorporating correct representations of the heart into a model trained from motion-free stacks, and in using it to perform motion correction. For instance, Oktay et al. \cite{Oktay2016} proposed to associate each SA slice with a set of probabilistic edge maps (PEMs) outlining the myocardial contours in the same slice as well as in the adjacent one, and to then perform rigid registration between the obtained PEMs.

\noindent \textbf{Contributions.} In this paper, we propose a comprehensive approach to automatically correct inter-slice respiratory motion in SA CMR image stacks. Our approach makes use of probabilistic segmentation maps (PSMs) of the left ventricular (LV) cavity generated with hybrid decision forests. PSMs are generated for each slice of the SA stack and rigidly registered in-plane to a target PSM. The main contributions of the paper are the following:
\begin{itemize}
\item The proposed approach includes two different techniques: if LA images are available, PSMs are generated from them and combined to create the target PSM. If not, the target PSM is produced from the same stack using a 3D model trained from motion-free stacks;
\item If LA images are available, the hybrid forests estimate from them at once both PSMs and landmarks locations for the apex and the mitral valve, which are used to restrict motion correction to the slices of the SA stack between them, thus limiting potential spurious results especially in the basal region;
\item The proposed approach was tested on a dataset acquired from 24 healthy subjects (for which anatomical 3D cardiac images were also available as reference) and compared to two techniques which use LA intensity images and LA segmentations (generated using FCNs) as targets, respectively. Testing was also performed after training the techniques on a different dataset to assess their generalisation properties.
\end{itemize}

\section{Methods}

\noindent \textbf{Hybrid Decision Forests.} A decision tree consists in the combination of split and leaf nodes arranged in a tree-like structure \cite{Criminisi2011}. Decision trees route a sample $\bm{x} \in \mathcal{X}$ (in our case an image patch) by recursively branching left or right at each split node $j$ until a leaf node $k$ is reached. Each leaf node is associated with a posterior distribution $p(y|\bm{x})$ for the output variable $y \in \mathcal{Y}$. Each split node $j$ is associated with a binary split function $h(\bm{x},\bm{\theta}_j) \in \{0,1\}$, defined by the set of parameters $\bm{\theta}_j$. During training, at each node the goal is to find the set of parameters $\bm{\theta}_j$ which maximizes a previously defined \emph{information gain} $I_j$, that is usually defined as $I_j = H(S_j) - \sum_{i \in \{0,1\}} \nicefrac{|S_j^i|}{|S_j|}\cdot H(S_j^i)$, where $S_j$, $S_j^0$ and $S_j^1$ are respectively the training set arriving at node $j$, leaving the node to the left and to the right. $H(S)$ is the entropy of the training set, whose construction depends on the task at hand (e.g. classification, regression). Different types of nodes (maximizing different information gains) can be interleaved within a single tree structure, thus called hybrid. In the present technique structured classification nodes (aiming at the generation of a PSM of the LV cavity) and regression nodes (aiming at landmark localisation) are combined \cite{Oktay2017}. During testing, the posterior distributions of the different trees are combined using an ensemble model.

Structured classification nodes associate to each image patch $\bm{x}$ a label $\bm{y} \in \mathcal{Y}$ consisting of a segmentation of the LV cavity within $\bm{x}$. Structured labels at each split node can be clustered into two subgroups depending on some similarity measure between them following a two-step procedure \cite{Dollar2015}. First, $\mathcal{Y}$ is mapped to an intermediate space $\mathcal{Z}$ by means of the function $\Pi: \mathcal{Y} \rightarrow \mathcal{Z}$ where the distance between labels can be computed. Then, PCA is applied to the vectors $\bm{z}$ to map the associated labels $\bm{y}$ into a binary set of labels $c \in \mathcal{C}$ $ = \{0,1\}$: this is achieved by applying a binary quantization to the principal component of each $\bm{z}$ vector. Finally, the Shannon entropy $H_{SC}(S) = - \sum_{c \in \mathcal{C}} p(c)log\big(p(c)\big)$ can be adopted, with $p(c)$ indicating the empirical distribution extracted from training set at the each node. Differently from \cite{Oktay2017}, in which edge maps were generated, to estimate segmentation maps we adopted the mapping
\begin{equation*}
\Pi: \bm{z} = [\bm{y}(j_1)=\bm{y}(j_2)=0] \oplus [\bm{y}(j_1)=\bm{y}(j_2)=1] \qquad \forall j_1 \neq j_2,
\label{eq_discretize_psm}
\end{equation*}

where $j_1$ and $j_2$ are indices spanning every pixel in $\bm{y}$. This mapping encodes for each pair of pixels in $\bm{y}$ whether they are both equal to 0 and whether they are both equal to 1, allowing the correct clustering of the labels at each node based on their similarity. At testing time, each sample patch of the test image is sent down each tree of the forest, and the segmentation maps stored at each selected leaf node are averaged producing a smooth segmentation map (PSM) of the LV cavity. The values in the PSM are proportional to the certainty in LV cavity detection, and can be used to assess the reliability of the prediction.

Regression nodes associate to each image patch $\bm{x}$ a label $\mathcal{D} = (\bm{d}^1, \bm{d}^2, \dotsc, \bm{d}^L)$, where $\bm{d}^l$ represents for each of the $L$ landmarks (LMs) the $N$-dimensional displacement vector from the patch centre to the landmark location \cite{Oktay2017}. The information gain used for regression nodes minimizes the determinant of the full covariance matrix $|\Lambda(S)|$ defined by the landmark displacement vectors: $H_{R}(S) = \frac{1}{2} log\big((2\pi e)^d|\Lambda(S)|$. The regression information is stored at each leaf node $k$ using a parametric model following a $N\cdot L$-dimensional multivariate normal distribution with $\overline{\bm{d}_k^l}$ and $\Sigma_k^l$ mean and covariance matrices, respectively. At testing time, for each landmark, Hough vote maps are generated by summing up the regression posterior distributions obtained from each tree for each patch \cite{Oktay2017}. Finally, the locations of the landmarks are determined by identifying the pixel with the highest value on each of the $L$ Hough vote maps. 

For training, the extracted features are multi-resolution image intensity, histogram of gradients (HoG) and gradient magnitude, exactly as in \cite{Oktay2017}. The described hybrid random forest approach is used to build several models: three for LA images (extracting at once PSMs and landmarks for the apex and the mitral valve) and two for the SA stacks (extracting 2D and 3D PSMs, respectively). These models are then used to perform motion correction with two different possible pipelines, depending on the availability of LA images.
\begin{figure}[t!]
\centering
\includegraphics[width=0.8\textwidth]{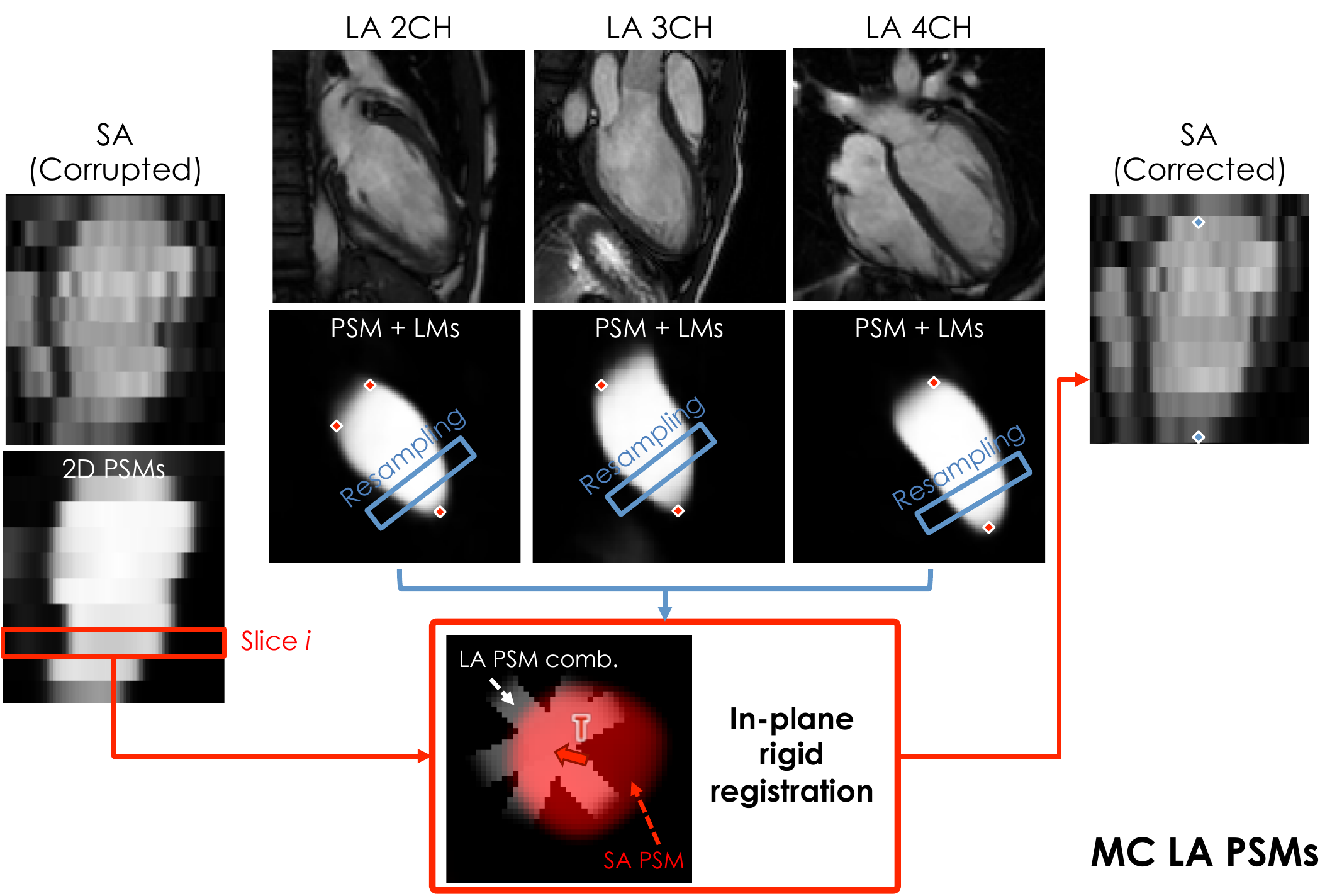}
\caption{Pipeline for motion correction using LA PSMs as target.}
\label{fig1}
\end{figure}

\noindent \textbf{Motion Correction with LA PSMs (MC LA PSMs).} This method relies on 2D SA PSMs generated from the motion-corrupted stack and on LA PSMs (together with landmarks), which are used as target (see Fig. \ref{fig1}). First, LA PSMs are rigidly registered (by 3D translation only, using normalized cross-correlation, NCC, as similarity metric) to the SA PSM stack to compensate for potential motion between different acquisitions. Then, for each slice of the SA PSM stack, the three registered LA PSMs are resampled and combined into a single image (referred to as LA PSM combined) containing the sections of the LA PSMs with respect to a specific slice. Finally, in-plane rigid registration (by translation only, using NCC) is performed between each SA PSM slice and the associated LA PSM combined, and the estimated translation is applied to the SA slice, thus performing the correction. These two steps (LA PSMs registration to the SA PSMs stack, and slice-by-slice SA PSMs registration to LA PSMs combined) are iterated until the maximum translation estimated within the stack is less than two pixels, which usually happens within the first 4 iterations. While this iterative registration scheme is similar to previously published ones \cite{Sinclair2017}, a major novelty is that not all the slices of the SA stack actually undergo motion correction: a slice is corrected only if a) its peak PSM value is above a threshold $T_m$ and b) it lies between the median apex and median mitral valve points (defined as medians of the landmark sets identified on the LA images). This allows the exclusion of slices outside the LV or with unreliable LV cavity detection.
\begin{figure}[t!]
\centering
\includegraphics[width=0.78\textwidth]{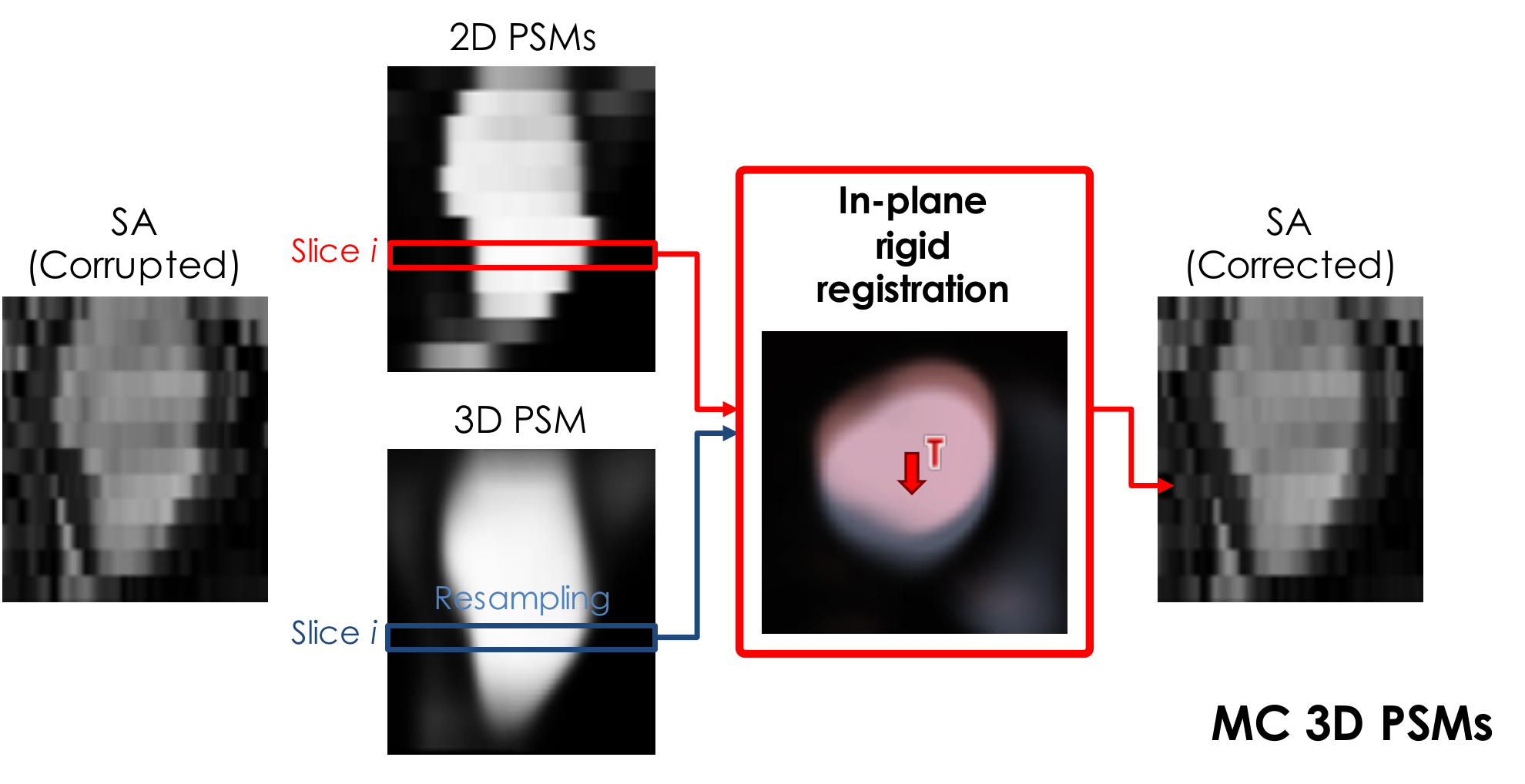}
\caption{Pipeline for motion correction using a 3D SA PSM as target.}
\label{fig2}
\end{figure}

\noindent \textbf{Motion Correction with 3D SA PSMs (MC 3D PSMs).} This method relies only on the information extracted from the motion-corrupted SA stack: 2D SA PSMs and a 3D SA PSM, which is used as target (see Fig. \ref{fig2}). While the models presented so far are trained using 2D patches $\bm{x}$ and labels $\bm{y}$, the 3D SA PSM one is trained using 3D patches encompassing 5 slices in the z direction. Training is performed on a set of high-resolution A3D images (inherently motion-free) with accompanying 3D segmentations, setting the patch thickness equal to that of 5 SA slices combined. This forces the model to learn representations of motion-free stacks. At testing, the model is applied to the SA stack (after an up-sampling step in the z direction to mimic the resolution of A3D images), generating a virtually motion-free 3D PSM which are used as a target for slice-by-slice in-plane registration (by translation only) of the 2D PSMs. The estimated translations are applied to the SA slices, thus performing the correction.

\section{Experiments and Results}

\noindent \textbf{Image Acquisition.} Two distinct CMR image datasets (obtained with different scanners, cardiac array coils and acquisition parameters) were used to test the proposed approach. The first dataset consists of 350 full CMR scans (including also A3D images) of healthy subjects, while the second one consists of 500 scans from the UK Biobank\footnote{http://www.ukbiobank.ac.uk/.}. Only end-diastolic frames were considered.
\begin{figure}[!t]
\centering
\includegraphics[width=\textwidth]{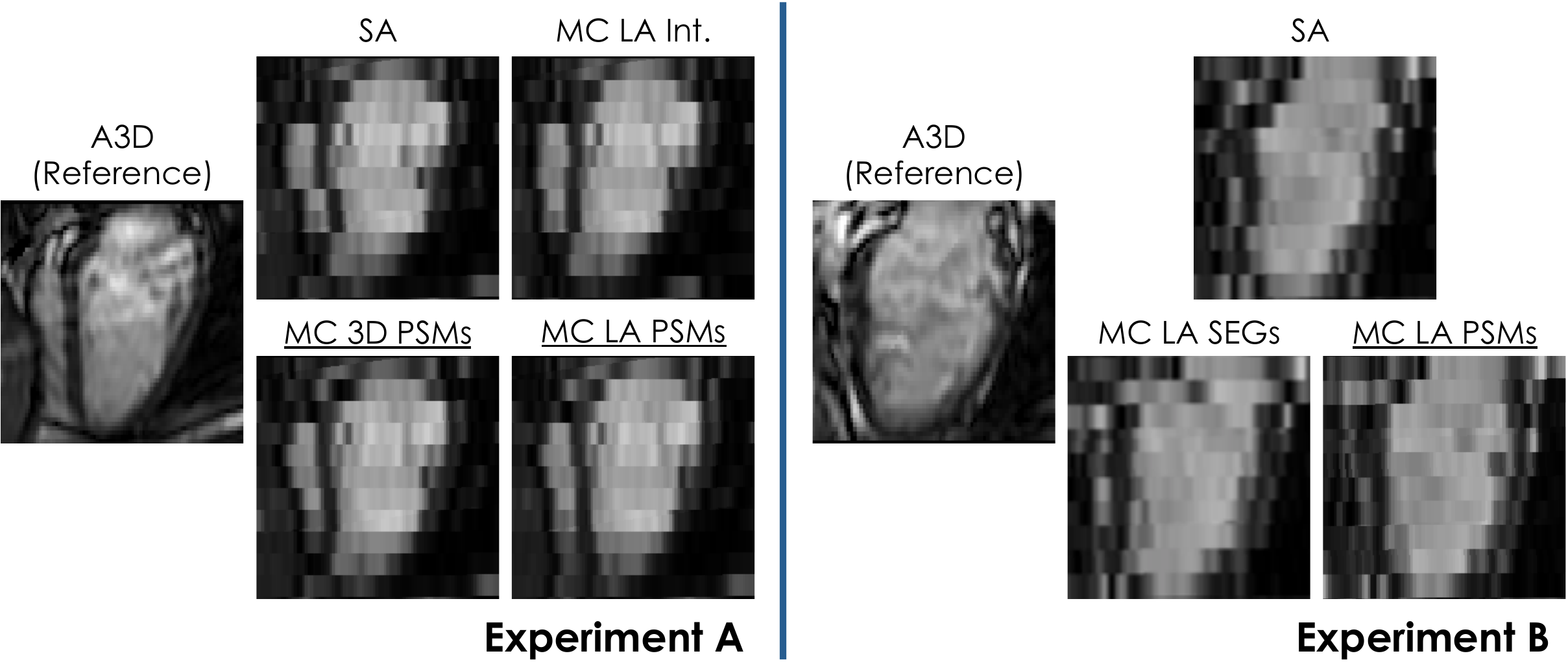}
\caption{Results obtained for two different subjects, one per experiment. The proposed techniques are underlined.}
\label{fig3}
\end{figure}

\noindent \textbf{Performance Evaluation.} Both datasets were annotated (either manually or automatically with subsequent manual corrections) to provide LV cavity segmentations for SA stacks, LA and A3D images as well as landmarks positions for the LA images alone. Two experiments were devised. For experiment A, 24 scans from the first dataset were extracted based on presence of visually-detected inter-slice motion and used as testing, while training was performed on the remaining scans of the same dataset. For this experiment, the proposed approach (with its two methods) was tested against an intensity-based technique (MC LA Int.) which iteratively registers SA slices to LA images (using normalized mutual information as similarity metric), essentially as in \cite{Lotjonen2004a}. For experiment B, testing was performed on the same 24 scans of experiment A, but training was performed on the whole second dataset. In this case, the proposed approach (using only MC LA PSMs) was tested against a technique (MC LA SEGs) which iteratively registers SA ``hard" segmentations to LA segmentations generated with FCNs (trained on images randomly extracted from the same database), essentially as in \cite{Sinclair2017}. To compare the accuracy of the implemented techniques, in both experiments the corrective translations estimated for each slice were applied to the provided SA segmentations. Then, the segmentation of the A3D images, considered as reference, were rigidly registered to the initial SA segmentation stack as well as to the those produced by each technique. Slice-by-slice in-plane registration was performed between reference and initial segmentation stack to identify motion corrupted slices, and those with more than 3 mm of misalignment were selected, for a total of 74. The evaluation of the accuracy of the implemented techniques was performed on these slices computing mean absolute distances (MAD), Hausdorff distances (HD) and Dice coefficients (DSC) between the LV cavity reference contours  and before or after motion correction. The average slice-by-slice relative improvements for each of these metrics were also computed, as well as the percentage of improved values (where 100\% would ideally mean that all of the corrupted slices improved their alignment). Of note, in experiment B, the testing set underwent histogram normalisation to match the intensity distributions of the training set. 

\noindent \textbf{Implementation Details.} For training, standard data augmentation was implemented (random rescaling following a normal distribution with mean 1 and std 0.1, random rotation following a normal distribution with mean 0\textdegree and std 30\textdegree). Image patch size was 48$\times$48 px for LA models and 32$\times$32 px for SA ones, segmentation label size 16$\times$16 px, number of samples 4$\cdot10^6$, number of trees 8. Finally, the threshold $T_m$ was set to 0.4 (on a scale from 0 to 1).
\begin{table}
\centering
\setlength\tabcolsep{6pt}
\begin{tabular}{r | c c | c c | c c | c}
\hline
\hline
\multirow{3}{1.8 cm}{\textbf{Motion Correction}} & \multicolumn{2}{c|}{\textbf{MAD}} & \multicolumn{2}{c|}{\textbf{HD}} & \multicolumn{2}{c|}{\textbf{DSC}} & \textbf{Ratio of}\\
& Mean & Mean & Mean & Mean & Mean & Mean & \textbf{improved}\\
& (mm) & Impr. & (mm) & Impr. & (a.u.) & Impr. & \textbf{slices}\\
\hline
\multicolumn{8}{c}{\textbf{Experiment A}}\\
\hline
None & 3.1 & & 6.9 & & 0.83 & &\\
MC LA Int. & 2.7 & 14\% & 6.2 & 12\% & 0.85 & 2\% & 77\%\\
\underline{MC 3D PSMs} & 2.6 & 19\% & 6.0 & 15\% & 0.86 & 2\% & 80\%\\
\underline{MC LA PSMs} & \textbf{1.9} & \textbf{38\%} & \textbf{4.9} & \textbf{29\%} & \textbf{0.89} & \textbf{7\%} & \textbf{92\%}\\
\hline
\multicolumn{8}{c}{\textbf{Experiment B}}\\
\hline
MC LA SEGs & 2.3 & 23\% & 5.7 & 16\% & 0.88 & \textbf{7\%} & 88\%\\
\underline{MC LA PSMs} & \textbf{2.1} & \textbf{33\%} & \textbf{5.2} & \textbf{26\%} & \textbf{0.89} & 6\% & \textbf{91\%}\\
\hline
\hline
\end{tabular}
\caption{Error metrics for experiments A (top) and experiment B (bottom). The proposed techniques are underlined.}
\label{tab1}
\end{table}
\vspace{-20pt}

\noindent \textbf{Results.} Approximate time to perform motion correction of one SA stack on a 6-core CPU is 25s for MC 3D PSMs and 36s for MC LA PSMs. The results for both experiment A and B are reported in Table \ref{tab1} and displayed for two cases in Fig. \ref{fig3}. Experiment A assesses the accuracy of the proposed approach in the scenario of training and testing performed in the same dataset. The results show that the the intensity-based method performs worse than the others, and that MC 3D PSMs obtains lower errors even without using LA images. MC LA PSMs is clearly the best method within this batch and is able to improve most (92\%) of the motion-corrupted slices. The fact that this method outperforms MC 3D PSMs was expected: while MC 3D PSMs can produce a smoothly aligned stack, more robustly than an intensity-based approach, it does not have any strong target for the realignment and relies only on the implicit model for the LV shape learned from the training set. Experiment B evaluates the accuracy of MC LA PSMs against a state-of-the-art approach like MC LA SEGs in a realistic scenario where motion correction has to be performed on a dataset completely different from the one used for training. Remarkably, MC LA PSMs produces results very similar to the ones obtained when trained on the same dataset. As expected given the similitudes between the two techniques, MC LA SEGs and MC LA PSMs perform similarly (no statistically significant differences were highlighted using a paired t-test). However, MC LA PSMs produces better mean results due to a higher robustness in the basal slices (see Fig. \ref{fig3}, right side): in fact, hard segmentation techniques have no safety mechanism to refrain from aligning slices in the basal region (usually beyond the actual basal slice) for which the FCN has produced spurious segmentations. On the contrary, the proposed approach has two: the check on the peak probability of the PSM and the comparison with the identified landmarks. As a result, the obtained motion correction tends to be more robust to this effect (see again Fig. \ref{fig3}, right side).

\section{Conclusion}

A comprehensive approach for fully-automated inter-slice motion correction for SA stacks has been presented. This approach relies on the generation of probabilistic segmentation maps of the LV cavity to drive slice-by-slice in-plane registration. It is able to handle cases in which no LA images are provided with a higher accuracy than common intensity-based methods that exploit them. When LA images are instead available, the proposed approach achieves results on par with methods based on hard segmentations while producing fewer outliers thanks to the simultaneous identification of landmarks to constrain the correction.

\subsubsection*{Acknowledgments.} This research has been conducted using the UK Biobank Resource under Application Number 18545. The first author benefited from a Marie Sklodowska-Curie Fellowship.

\bibliographystyle{splncs}
\bibliography{bibliography}

\end{document}